\def\year{2019}\relax
\documentclass[letterpaper]{article} 
\usepackage{aaai19}  
\usepackage{times}  
\usepackage{helvet}  
\usepackage{courier}  
\usepackage{url}  
\usepackage{graphicx}  

\usepackage{natbib}
\usepackage{amsmath,amsfonts,amssymb, bm}
\DeclareMathOperator*{\argmax}{argmax}
\DeclareMathOperator*{\argmin}{argmin}
\usepackage[ruled,vlined]{algorithm2e}

\begin{document}

\title{Dependency Grammar Induction \\with a Neural Variational Transition-based Parser}
\author{
Bowen Li, Jianpeng Cheng, Yang Liu \and Frank Keller \\
Institute for Language, Cognition and Computation \\
School of Informatics, University of Edinburgh \\ 
10 Crichton Street, Edinburgh EH8 9AB, UK \\
{\tt \{bowen.li, jianpeng.cheng, yang.liu2\}@ed.ac.uk, keller@inf.ed.ac.uk} \\
}
\maketitle
\begin{abstract}
  Dependency grammar induction is the task of learning dependency
  syntax without annotated training data.  
  Traditional graph-based models with global inference achieve state-of-the-art results on
  this task but they require $O(n^3)$ run time. 
  Transition-based models enable faster inference with $O(n)$ time complexity, but their performance still lags behind. 
  In this work, we propose a neural transition-based parser for dependency grammar induction,
  whose inference procedure utilizes rich neural features with $O(n)$ time complexity. 
  We train the parser with an integration of variational inference, posterior regularization and variance reduction techniques. 
  The resulting framework outperforms previous unsupervised transition-based dependency parsers and achieves performance comparable to graph-based models, both on the English Penn Treebank
  and on the Universal Dependency Treebank. 
  In an empirical comparison, we show that our approach substantially increases parsing speed over graph-based models.
\end{abstract}

\section{Introduction}
\label{sec:intro}

Grammar induction is the task of deriving plausible syntactic
structures from raw text, without the use of annotated training data.
In the case of dependency parsing, the syntactic structure takes the
form of a tree whose nodes are the words of the sentence, and whose
arcs are directed and denote head-dependent relationships between
words.  Inducing such a tree without annotated training data is
challenging because of data sparseness and ambiguity, and because the
search space of potential trees is huge, making optimization
difficult.

Most existing approaches to dependency grammar induction have used
inference over graph structures and are based either on the dependency
model with valence (DMV) of \citet{klein2004corpus} or the maximum
spanning tree algorithm (MST) for dependency parsing by
\citet{mcdonald2011multi}.
State-of-the-art representatives include LC-DMV \citep{noji2016using}
and Convex-MST \citep{grave2015convex}.  Recently, researchers have
also introduced neural networks for feature extraction in graph-based
models \citep{jiang2016unsupervised, cai2017crf}.

Though graph-based models
 achieve impressive results, their inference procedure requires $O(n^3)$ time complexity. 
Meanwhile, features in graph-based models must be decomposable over substructures to enable dynamic programming.
In comparison, transition-based models allow faster inference with
linear time complexity and richer
feature sets. Although relying on local inference,
transition-based models have been shown to perform well in supervised
parsing \citep{simplelstmdep2016, dyer2015transition}.  However, unsupervised
transition parsers are not well-studied.  One
exception is the work of \citet{rasooli2012fast}, in which
search-based structure prediction \citep{daume2009unsupervised} is
used with a simple feature set.  However, there is still a large
performance gap compared to graph-based models.

Recently, \citet{dyer2016recurrent} proposed recurrent neural
network grammars (RNNGs)---a probabilistic transition-based model for 
constituency trees.  RNNG can be used either in a generative way as a
language model or in a discriminative way as a parser.
\citet{cheng2017generative}
 use an autoencoder to integrate
discriminative and generative RNNGs, yielding a reconstruction process
with parse trees as latent variables and enabling the two components
to be trained jointly on a language modeling objective.  However, their work uses
observed trees for training and does not study unsupervised learning.

In this paper, we make a more radical departure from the existing
literature in dependency grammar induction, by proposing an
unsupervised neural variational transition-based parser.
Specifically, we first modify the transition actions in the original
RNNG into a set of arc-standard actions for projective dependency
parsing, and then build a dependency variant of the model of
\citet{cheng2017generative}.  Although this approach performs well for
supervised parsing, when applied in an unsupervised setting, the
performance decreases dramatically (see 
Experiments for details).  We hypothesize that this is because the parser is fairly unconstrained without prior
linguistic knowledge
\citep{naseem2010using, noji2016using}.  Therefore, we augment the
model with posterior regularization, allowing us to seamlessly
integrate linguistic knowledge in the shape of a small number of
universal linguistic rules.  In addition, we propose a novel variance
reduction method for stabilizing neural variational inference with
discrete latent variables. This yields the first known model that
makes it possible to use posterior regularization for neural variational inference with
discrete latent variables.
 When evaluating on the English Penn Treebank and
on eight languages from the Universal Dependency (UD) Treebank, we
find that our model with posterior regularization outperforms the best
unsupervised transition-based dependency parser
\citep{rasooli2012fast}, and approaches the performance of graph-based
models.  We also show how a weak form of supervision can be integrated
elegantly into our framework in the form of rule
expectations. Finally, we present empirical evidence for the
complexity advantage of transition-based models: our model attains a
large speed-up compared to a state-of-the-art graph-based model. Code
and Supplementary Material are available.\footnote{\tt
  https://github.com/libowen2121/\\VI-dependency-syntax}

\section{Background}
\label{sec:background}

RNNG is a top-down transition system originally proposed for
constituency parsing and generation.  There are two variants: the
discriminative RNNG and the generative RNNG.  The discriminative RNNG
takes a sentence as input, and predicts the probability of
generating a corresponding parse tree from the sentence. The model uses a buffer to
store unprocessed terminal words and a stack to store partially
completed syntactic constituents.  It then follows top-down transition
actions to shift words from the buffer to the stack to construct
syntactic constituents incrementally.

The discriminative RNNG can be modified slightly to formulate the
generative RNNG, an algorithm for incrementally producing trees and
sentences in a generative fashion.  In generative RNNG, there is no buffer of unprocessed
words, but there is an output buffer for storing words that have been
generated.  Top-down actions are then specified to generate words and
tree non-terminals in pre-order.  Though not able to parse on its own,
a generative RNNG can be used for language modeling as long as parse
trees are sampled from a known distribution.

We modify the transition actions in the original RNNG into a set of
arc-standard actions for projective dependency parsing.
In the discriminative modeling case, the action space includes:
\begin{itemize}
	\setlength{\parskip}{0pt}
	\setlength{\itemsep}{0pt plus 1pt}
	\item \textsc{shift} fetches the first word in the buffer and
	pushes it onto the top of the stack.
	\item \textsc{left-reduce} adds a left arc in between the top two
	words of the stack and merges them into a single construct.  
	\item \textsc{right-reduce} adds a right arc in between the top two
	words of the stack and merges them into a single construct.
\end{itemize}
In the generative modeling case, the \textsc{shift} operation is
replaced by a \textsc{gen} operation:
\begin{itemize}
	\item \textsc{gen} generates a word and adds it to the stack and the
	output buffer.
\end{itemize}
%

\section{Methodology}
\label{sec:methods}

To build our dependency grammar induction model, we follow \citet{cheng2017generative} and propose
a dependency-based, encoder-decoder RNNG.  This model includes
(1) a
discriminative RNNG as the
\emph{encoder} 
for mapping the input sentence into a latent variable, which for the
grammar induction task is a sequence of parse actions for building the
dependency tree; (2) a generative RNNG as the
\emph{decoder} 
for reconstructing the input sentence based on the latent parse
actions. The training objective is the likelihood of the observed input sentence, which is reformulated as an evidence lower bound
(ELBO), and solved with neural variational inference.
The REINFORCE algorithm \citep{williams1992simple} is utilized to handle discrete latent variables
in optimization. 
Overall, the encoder and decoder are jointly trained,
inducing  latent parse trees or actions from only unlabelled text data.  To
further regularize the space of parse trees with
a linguistic prior, we introduce posterior regularization into the
basic framework. 
Finally, we propose a novel variance reduction technique to train our
posterior regularized framework more effectively.

\subsection{Encoder}
\label{chap:Encoder}

We formulate the encoder as a discriminative dependency RNNG that
computes the conditional probability $p(a | x)$ of the transition
action sequence $a$ given the observed sentence $x$.  The conditional
probability is factorized over time steps, and parameterized by 
a transitional state embedding~$v$:
\begin{equation}
\label{eq:encoder}
p(a | x) = \prod_{t=1}^{|a|} p(a_t | v_t)
\end{equation}
where $v_t$ is the transitional state embedding of the encoder at time
step~$t$.
The encoder is the actual component for \textbf{parsing} at run time.

\subsection{Decoder}
\label{chap:Decoder}

The decoder is a generative dependency RNNG that models the joint
probability $p(x, a)$ of a latent transition action sequence $a$ and an
observed sentence $x$. 
This joint distribution can be factorized into a sequence of action and 
word (emitted by \textsc{gen}) probabilities, which are parameterized by 
a transitional state embedding~$u$:
\begin{equation}
\begin{split}
p(x, a) & = p(a)p(x | a) \\
& = \prod_{t=1}^{|a|} p(a_t | u_t)p(x_t | u_t)^{I(a_t={\textsc{gen})}}
\end{split}
\end{equation}
where $I$ is an indicator function and $u_t$ is the state embedding at
time step $t$. The features and the modeling details of both the
encoder and the decoder can be found in the Supplementary Material.

\subsection{Training Objective}
\label{chap:Training}

Consider a latent variable model in which the encoder infers the
latent transition actions (i.e.,~the dependency structure) and the
decoder reconstructs the sentence from these actions. The maximum
likelihood estimate of the model parameters is determined by the log
marginal likelihood of the sentence:
\begin{equation}
\log p(x) = \log\sum_{a} p(x, a)
\end{equation}
Since the form of the log likelihood is intractable in our case,  we optimize the ELBO (by
Jensen's Inequality) as follows:
\begin{equation}
\label{eq:loss}
\begin{split}
\log p(x) & \geqslant \log p(x) - KL[q(a) || p(a|x)] \\
&  = \mathbb{E}_{q(a)} [ \log \frac{p(x, a)}{q(a)} ] = \mathcal{L}_x
\end{split}
\end{equation}
where $KL$ is the Kullback-Leibler divergence and $q(a)$ is the
variational approximation of the true posterior. This training objective is optimized with the EM algorithm.  In the E-step, we
approximate the variational distribution $q(a)$ based on the encoder
and the observation $x$---$q(a)$ is parameterized as $q_{\omega}(a | x)$.
 Similarly, the joint probability $p(x, a)$ is parameterized
by the decoder as $p_{\theta}(x, a)$.

In the M-step, the decoder parameters $\theta$ can be directly
updated by gradient descent via Monte Carlo simulation:
\begin{equation}
\begin{split}
\frac{\partial \mathcal{L}_x}{\partial \theta} & = \mathbb{E}_{q_\omega(a | x)} [ \frac{\partial \log p_\theta (x, a)}{\partial \theta} ] \\
& \approx \frac{1}{M} \sum_m \frac{\partial \log p_\theta (x, a^{(m)})}{\partial \theta}
\end{split}
\end{equation}
where $M$ samples $a^{(m)} \sim q_\omega(a | x)$ are drawn
independently to compute the stochastic gradient.

For the encoder parameters $\omega$, since the sampling operation is
not differentiable, we approximate the gradients using the REINFORCE
algorithm \citep{williams1992simple}:
%
\begin{equation}
\begin{split}
\label{eq: rl}
\frac{\partial \mathcal{L}_x}{\partial \omega} & = \mathbb{E}_{q_\omega(a | x)} [ l(x, a) \frac{\partial \log q_\omega(a | x)}{\partial \omega} ] \\
& \approx \frac{1}{M} \sum_m l(x, a^{(m)}) \frac{\partial \log q_\omega (a^{(m)} | x)}{\partial \omega}
\end{split}
\end{equation}
where $l$ is known as the score function and computed as:
\begin{equation}
  \label{eq:original score}
  l(x,a) = \log \frac{p_{\theta}(x, a)} {q_\omega(a|x)}
\end{equation}

\subsection{Posterior Regularization}
\label{chap:Training with PR}

As will become clear in the 
Experiments section,
the basic model discussed previously
performs poorly when used for unsupervised parsing, barely
outperforming a left-branching baseline for English.  We hypothesize
the reason is that the basic model is fairly unconstrained: without any constraints to
regularize the latent space, the induced parses will be arbitrary,
since the model is only trained to maximize sentence likelihood
\citep{naseem2010using, noji2016using}.

We therefore introduce posterior regularization (PR;
\citealt{ganchev2010posterior}) to encourage the neural network to
generate well-formed trees.  Via posterior regularization, we can give
the model access to a small amount of linguistic information in the
form of universal syntactic rules \citep{naseem2010using}, which are
the same for all languages.  These rules effectively function as
features, which impose soft constraints on the neural parameters in
the form of expectations.

To integrate the PR constraints into the model, a set $Q$ of allowed
posterior distributions over the hidden variables $a$ can be defined
as:
\begin{equation}
\label{eq:PR set}
\begin{split}
Q = \{q(a): \exists\bm{\xi},~\mathbb{E}_q[\bm{\phi}(x,a)] - \bm{b} \leqslant \bm{\xi};~||\bm{\xi}||_{\beta}\leqslant\varepsilon \}
\end{split}
\end{equation} 
where $\bm{\phi}(x,a)$ is a vector of feature functions, $\bm{b}$ is a
vector of given negative expectations, $\bm{\xi}$ is a vector of slack
variables, $\varepsilon$ is a predefined small value and
$||\cdot||_{\beta}$ denotes some norm.  The PR algorithm only works if
$Q$ is non-empty.

In dependency grammar induction, $\phi_k(x,a)$ (the $k^{th}$ element
in $\bm{\phi}(x,a)$) can
be set as the negative number of times a given rule
(dependency arcs, e.g.,~\textit{Root $\to$ Verb}, \textit{Verb $\to$ Noun}) occurs in a
sentence. We hope to bias the learning so that each sentence is parsed
to contain these kinds of arcs more than a threshold in the
expectation.  The posterior regularized likelihood is then:
\begin{equation}
\label{eq:PR loss}
\begin{split}
\mathcal{L}_Q & = \max_{q \in Q} \mathcal{L}_x \\
& = \log p(x) - \min_{q \in Q} KL[q(a) ~||~ p(a|x)]
\end{split}
\end{equation} 
Equation~(\ref{eq:PR loss}) indicates that, in the posterior
regularized framework, $q(a)$ not only approximates the true posterior
$p(a|x)$ (estimated by the encoder network $q_{\omega}(a|x)$) but also
belongs to the constrained set $Q$.  To optimize $\mathcal{L}_Q$ via
the EM algorithm, we get the revised E$^\prime$-step as:
\begin{equation}
\begin{split}
q(a)  & = \argmax_{q \in Q} \mathcal{L}_Q \\
& = \argmin_{q \in Q} KL[q(a) ~||~ q_\omega(a|x)]
\end{split}
\end{equation} 
Formally, the optimization problem in the E$^\prime$-step can be described as:
\begin{equation}
\label{eq:PR optim}
\begin{split}
\min_{q,\bm{\xi}} & ~~~KL[q(a) ~||~ q_{\omega}(a|x)] \\
s.t. & ~~~\mathbb{E}_q[\bm{\phi}(x,a)] - \bm{b} \leqslant \bm{\xi}; ~~||\bm{\xi}||_{\beta}\leqslant\varepsilon
\end{split}
\end{equation} 
Following \citet{ganchev2010posterior}, we can solve the optimization
problem in~(\ref{eq:PR optim}) in its Lagrangian dual form. Since our
transition-based encoder satisfies the decomposition property, the
conditional probability $q_{\omega}(a|x)$ can be factored as
$\prod_{t=1}^{|a|} q_{\omega}(a_t | v_t)$ in~(\ref{eq:encoder}). Thus,
the factored primal solution can be written as:
\begin{equation}
\label{eq: pr_q}
\begin{split}
q(a) =  \frac{q_{\omega}(a|x)}{Z(\bm{\lambda}^{*})} \exp(-\bm{\lambda}^{*T}\bm{\phi}(x,a))
\end{split}
\end{equation}
where $\bm{\lambda}$ is the Lagrangian multiplier whose solution is
given as
$\bm{\lambda}^* = \argmax_{\bm{\lambda}\geqslant 0}
-\bm{b}^T\bm{\lambda} - \log Z(\bm{\lambda}) - \varepsilon
||\bm{\lambda}||_{\beta^*}$\footnote{$||\cdot||_{\beta^*}$
  is the dual norm of $||\cdot||_{\beta}$. Here we use $\ell_2$ norm
  for both primal norm $||\cdot||_{\beta}$ and dual norm
  $||\cdot||_{\beta^*}$. } and $Z(\bm{\lambda})$ is given as:
\begin{equation}
Z(\bm{\lambda})=\sum_a q_{\omega}(a | x) \exp(-\bm{\lambda}^T\bm{\phi}(x,a))
\end{equation}
We also define the multiplier computed by PR as:
\begin{equation}
\label{eq:gamma}
\gamma(a, x)=\frac{1}{Z(\bm{\lambda})} \exp(-\bm{\lambda}^{T} \bm{\phi}(x,a))
\end{equation}
In our case, computing the normalization term $Z(\bm{\lambda})$ is
intractable for transition-based dependency parsing systems. To
address this problem, we view $Z(\bm{\lambda})$ as an expectation and
estimate it by Monte Carlo simulation as:
\begin{equation}
\begin{split}
Z(\bm{\lambda}) &= \mathbb{E}_{q_{\omega}(a|x)}[\exp(-\bm{\lambda}^T\bm{\phi}(x,a))] \\
 &\approx \frac{1}{M} \sum_m \exp(-\bm{\lambda}^T\bm{\phi}(x,a^{(m)}))
 \end{split}
\end{equation}
Finally, we compute the gradients for encoder and decoder in the
M-step as follows:
\begin{equation}
  \begin{split}
    \frac{\partial \mathcal{L}_x}{\partial \theta} 
    &= \frac{1}{M} \sum_m  \gamma(x, a^{(m)}) l(x, a^{(m)}) \frac{\partial \log p_\theta (x, a^{(m)})}{\partial \theta} \\
    \frac{\partial \mathcal{L}_x}{\partial \omega} 
    &= \frac{1}{M} \sum_m  \gamma(x, a^{(m)}) l(x, a^{(m)}) \frac{\partial \log q (a^{(m)})}{\partial \omega}
  \end{split}
  \label{pr update}
\end{equation}
where $l$ is the score function computed as in~(\ref{eq:original
  score}). Details of the derivation of the M-step can be found in the
Supplementary Material.

\subsection{Variance Reduction in the M-step}
\label{chap:Reduction}

Training a neural variational inference framework with discrete latent
variables is known to be a challenging problem
\citep{nvil2014,miao2016language,miao2016neural}.  This is mainly
caused by the sampling step of discrete latent variables which results
in high variance, especially at the early stage of training when both
encoder and decoder parameters are far from optimal.  Intuitively, the
score function $l(x,a)$ weights the gradient for each latent sample
$a$, and its variance plays a crucial role in updating the parameters
in the M-step.

To reduce the variance of the score function and stabilize learning,
previous work \citep{nvil2014,miao2016language,miao2016neural} adopts
the \emph{baseline method} (\textsc{RL-BL}), re-defining the score
function as:
\begin{equation}
l_{\textsc{RL-BL}}(x, a) = l(x,a) - b(x) - b
\end{equation} 
where $b(x)$ is a parameterized, input-dependent baseline (e.g.,~a
neural language model in our case) and $b$ is the bias. The baseline
method is able to reduce the variance to some extent, but also
introduces extra model parameters that complicate optimization. In the
following we propose an alternative generic method for reducing the
variance of the gradient estimator in the M-step, as well as another
task-specific method which results in further improvement.

\paragraph{1. Generic Method} The intuition behind the generic method is
as follows: the algorithm takes $M$ latent samples for each input $x$
and a score $l(x, a^{(m)})$ is computed for each sample $a^{(m)}$,
hence the variance can be reduced by normalization within the group of
samples. This motivates the following normalized score function
$l_{\textsc{RL-SN}}(x,a)$:
\begin{equation}
l_{\text{RL-SN}}(x,a) = \frac{l(x,a) - \bar{l}(x,a)}{\max(1, \sqrt{\mathrm{Var}[l(x,a)]})}
\end{equation}

\paragraph{2. Task-Specific Method}
Besides the generic variance reduction method which applies to
discrete neural variational inference in general, we further propose
to enhance the quality of the score function $l_{\textsc{RL-SN}}(x,a)$
for the specific dependency grammar induction task.

Intuitively, the score function in~(\ref{pr update}) weights the
gradient of a given sample $a$ by a positive or negative value, while
$\gamma(x, a)$ only weights the gradient by a positive value.  As a
result, the score function plays a much more significant role in
determining the optimization direction.  Therefore, we propose to
correct the polarity of our $l_{\textsc{RL-SN}}(x,a)$ with the number
of rules $s(x,a) = -\textsc{sum}[\bm{\phi}(x,a)]$ that occur in the
induced dependency structure, where $\textsc{sum}[]$ returns the sum
of vector elements. The refined score function is:
\begin{equation}
\label{eq:rl-sn}
\begin{split}
l_{\textsc{RL-PC}}(x,a)= 
&\begin{cases}
~\,\,|l_{\textsc{RL-SN}}(x,a)|& \hat{s}(x,a) \geqslant 0 \\
-|l_{\textsc{RL-SN}}(x,a)|& \hat{s}(x,a) < 0
\end{cases}
\end{split}
\end{equation}
where $\hat{s}(x,a) = \frac{s(x,a) - \bar{s}(x,a)}{\sqrt{\mathrm{Var}[s]}}$.

Since $\hat{s}(x,a)$ provides a natural corrective, 
we can obtain a simpler variant of~(\ref{eq:rl-sn}) by
directly using $\hat{s}(x,a)$ as the score function:
\begin{equation}
l_{\textsc{RL-C}}(x,a)= \hat{s}(x,a)
\end{equation}
We will experimentally compare the different variance reduction
techniques (or score functions) of the reinforcement learning
objective.

\section{Experiments}
\label{sec:experiments}

\subsection{Datasets, Universal Rules, and Setup}

\paragraph{English Penn Treebank}
We use the Wall Street Journal (WSJ) section of the English Penn
Treebank \citep{marcus1993building}.  The dataset is preprocessed to
strip off punctuation.  We train our model on sections 2--21, tune
the hyperparameters on section 22, and evaluate on section 23.
Sentences of length $\leq 10$ are used for training, and we report
directed dependency accuracy (DDA) on test sentences of length
$\leq 10$ (WSJ-10), and on all sentences (WSJ).

\paragraph{Universal Dependency Treebank}
We select eight languages from the Universal Dependency Treebank 1.4
\citep{nivre2016universal}. We train our model on training sentences
of length $\leq 10$ and report DDA on test sentences of length
$\leq 15$ and $\leq 40$.  We found that training on short sentences
generally increased performance compared to training on longer
sentences (e.g.,~length $\leq 15$).

\paragraph{Universal Rules}
We employ the universal linguistic rules of \citet{naseem2010using}
and \citet{noji2016using} for WSJ and the Universal Dependency
Treebank, respectively (details can be found in the Supplementary
Material).  For WSJ, we expand the coarse rules defined in
\citet{naseem2010using} with the Penn Treebank fine-grained
part-of-speech tags.  For example, \textit{Verb} is expanded as
\textit{VB}, \textit{VBD}, \textit{VBG}, \textit{VBN}, \textit{VBP}
and \textit{VBZ}.

\paragraph{Setup}
To avoid a scenario in which REINFORCE has to work with an arbitrarily
initialized encoder and decoder, our posterior regularized neural
variational dependency parser is pretrained with the direct reward
from PR. (This will be discussed later; for more details on the
training, see Supplementary Material.)

We use AdaGrad \citep{duchi2011adaptive} to optimize the parameters
of the encoder and decoder, as well as the projected gradient descent
algorithm \citep{bertsekas1999nonlinear} to optimize the parameters of
posterior regularization.

We use GloVe embeddings \citep{pennington2014glove} to initialize
English word vectors and FastText embeddings
\citep{bojanowski2016enriching} for the other languages.  Across all
experiments, we test both unlexicalized and lexicalized versions of
our models.  The unlexicalized versions use gold POS tags as model
inputs, while the lexicalized versions additionally use word tokens
\citep{le2015unsupervised}.  We use Brown clustering
\citep{brown1992class} to obtain additional features in the
lexicalized versions~\citep{buys2015generative}.

We report average DDA and best DDA over five runs for our main parsing
results.

\subsection{Exploration of Model Variants}
\label{chap:exploration}

\paragraph{Posterior Regularization}
To study the effectiveness of posterior regularization in the neural
grammar induction model, we first implement a fully unsupervised model
without posterior regularization.
This model is trained with variational inference, using the
standard REINFORCE objective with a \emph{baseline}
\citep{nvil2014,miao2016language,miao2016neural} and employing no
posterior regularization.

Table~\ref{tab:vanilla wsj} shows the results for the unsupervised
model, together with the random and left- and right-branching
baselines.  We observe that the unsupervised model (both the
unlexicalized and lexicalized versions) fails to beat the
left-branching baseline.  These results suggest that without any prior
linguistic knowledge, the trained model is fairly unconstrained.  A
comparison with posterior-regularized results in
Table~\ref{tab:pretraining wsj} (to be discussed next) reveals the
effectiveness of posterior regularization in incorporating such
knowledge.

\begin{table}[tb]
  \centering
  \begin{tabular}{ l  c  c}
    \noalign{\hrule height 1pt}
    Model                      & WSJ-10   & WSJ \\  
    \noalign{\hrule height 1pt}
    Random                  & 19.1    & 16.4  \\
    Left branching          & 36.2    & 30.2  \\  
    Right branching         & 20.1    & 20.6  \\
    \hline
    \textsc{Unsupervised}     & 33.3 (39.0)  & 29.0 (30.5) \\
    L-\textsc{Unsupervised}   & 34.9 (36.4)  & 28.0 (30.2) \\
    \noalign{\hrule height 1pt}
  \end{tabular}
  \caption{Evaluation of the fully unsupervised model (without
    posterior regularization) on the English Penn Treebank. We report 
    average DDA and the best DDA (in brackets) over five runs. ``L-''
    denotes the lexicalized version.}
  \label{tab:vanilla wsj}
\end{table}

\paragraph{Pretraining}
Unsupervised models in general face a \textit{cold-start} problem
since no gold annotations exist to ``warm up'' the model parameters
quickly.  This can be observed in~(\ref{pr update}): the gradient
updates of the model are dependent on the score function $l$, which in
return relies on the model parameters.  At the beginning of training
we cannot obtain an accurately approximated $l$ for updating model
parameters.  To alleviate this problem, one approach is to ignore the
score function in the gradient update at the early stage.  In this
case, both the encoder and decoder are trained with the direct reward
from PR (detailed equations can be found in the Supplementary
Material).  We test the effectiveness of this approach, which we call
\textit{pretraining}.

Table \ref{tab:pretraining wsj} shows the results of a standard
posterior-regularized model compared to one only with pretraining.  
Both models use the unlexicalized setup.  We find that the
posterior-regularized model benefits a lot from pretraining, which
therefore is a useful way to avoid cold start.

\begin{table}[tb]
  \centering
  \begin{tabular}{ l  c  c}
    \noalign{\hrule height 1pt}
                       & WSJ-10        & WSJ \\  
    \noalign{\hrule height 1pt}
    No Pretraining          & 47.5 (59.8)   & 36.7 (46.3)  \\
    Pretraining             & 64.8 (67.1)   & 42.0 (43.7) \\
    \noalign{\hrule height 1pt}
  \end{tabular}
  \caption{Evaluation of the posterior-regularized model with and
    without pretraining on the WSJ. We report average DDA and best DDA
    (in brackets) over five runs.}
  \label{tab:pretraining wsj}
\end{table}

\begin{table}[tb]
  \centering
  \begin{tabular}{ c c c c c}
    \noalign{\hrule height 1pt}
      &  \textsc{RL-BL} & \textsc{RL-SN} & \textsc{RL-C} & \textsc{RL-PC} \\
    \noalign{\hrule height 1pt}
    $\mu$     &  58.7   &  60.8  &   64.4  & 66.7  \\
    $\sigma$  &  1.8    &  0.6   &  0.3    & 0.7   \\
    \noalign{\hrule height 1pt}
  \end{tabular}
  \caption{Comparison of models with different variance reduction
    techniques (or score functions) on the WSJ-10 test set. We 
    report the average DDA $\mu$ and its standard deviation $\sigma$ over five runs.}
  \label{tab:critic}
\end{table}

\paragraph{Variance Reduction}
Previously, we described various variance reduction techniques, or
modified score functions, for the reinforcement learning objective.
These include the conventional \textit{baseline} method
(\textsc{RL-BL}), our sample normalization method (\textsc{RL-SN}),
sample normalization with additional polarity correction
(\textsc{RL-PC}), and a simplified version of the later
(\textsc{RL-C}).  We now compare these techniques; all experiments
were conducted with pretraining and on the unlexicalized model.

The experimental results in Table~\ref{tab:critic} show that
\textsc{RL-SN} outperforms \textsc{RL-BL} on average DDA, which
indicates that sample normalization is more effective in reducing the
variance of the gradient estimator.  We believe the gain comes from
the fact that sample normalization does not introduce extra model
parameters, whereas \textsc{RL-BL} does.  Polarity correction further
boosts performance. However, polarity correction uses the number of
universal rules present in a induced dependency structure, i.e., it is
a task-specific method for variance reduction. Also \textsc{RL-C} (the
simplified version of \textsc{RL-PC}) achieves competitive
performance.

\paragraph{Universal Rules}
In our PR scheme, the rule expectations can be uniformly
initialized. This approach does not require any annotated training
data; the parser is furnished only with a small set of universal
linguistic rules. We call this setting \textsc{UniversalRules}.

However, we can initialize the rule expectation
non-uniformly, which allows us to introduce a degree of supervision
into the PR scheme. Here, we explore one way of doing this: we assume
a training set that is annotated with dependency rules (the training
portion of the WSJ), based on which we estimate expectations for the
universal rules. We call this setting \textsc{WeaklySupervised}.

The results of an experiment comparing these two settings is shown in
Table~\ref{tab:rules wsj}. In both cases we use pretraining and the
best performing score function \textsc{RL-PC}. Here we report results
using both unlexicalized and lexicalized settings.  It can be seen
that the best performing \textsc{UniversalRules} model is the
unlexicalized one, while the best \textsc{WeaklySupervised} model is
lexicalized.  Overall, \textsc{WeaklySupervised} outperforms
\textsc{UniversalRules}, which demonstrates that our posterior
regularized parser is able to effectively use weak supervision in the
form of an empirical initialization of the rule expectations.

\begin{table}[tb]
  \centering
  \scalebox{1.0}{\begin{tabular}{ l  c  c}
    \noalign{\hrule height 1pt}
    Model                      & WSJ-10   & WSJ \\             
    \noalign{\hrule height 1pt}
    \textsc{UniversalRules}          &  \textbf{54.7} (58.2)          &  \textbf{37.8} (39.3)  \\
    L-\textsc{UniversalRules}      &  54.7 (56.3)          &  36.8 (38.1)   \\
    \hline
    \textsc{WeaklySupervised}     &  66.7 (67.6)          &  43.6 (45.0)   \\
    L-\textsc{WeaklySupervised}   &  \textbf{68.2} (71.1)          &  \textbf{48.6} (50.2)   \\
    \noalign{\hrule height 1pt}
  \end{tabular}}
  \caption{Comparison of uniformly initialized
    (\textsc{UniversalRules}) and empirically estimated 
    (\textsc{WeaklySupervised}) rule expectation on the WSJ.}
  \label{tab:rules wsj}
\end{table}

\subsection{Parsing Results}
\label{sec:res}

\begin{table}[tb]
  \centering
  \scalebox{1.0}{\begin{tabular}{ l  c  c}
    \noalign{\hrule height 1.5pt}
    Model                   & WSJ-10   & WSJ \\ 
    \noalign{\hrule height 1pt}
    \multicolumn{3}{l}{Graph-based models} \\
    \hline
    Convex-MST            & 60.8    & \textbf{48.6}  \\      
    HDP-DEP               & \textbf{71.9}    & --    \\ 
    \noalign{\hrule height 1pt}
    \multicolumn{3}{l}{Transition-based models} \\
    \hline
    RF                     & 37.3 (40.7)    & 32.1 (33.1) \\
    RF+H1+H2           & 51.0 (52.7)    & 37.2 (37.6) \\  
    \textsc{UniversalRules}   &  54.7 (58.2)          &  37.8 (39.3)  \\
    L-\textsc{WeaklySupervised}            &  68.2 (71.1) & \textbf{48.6} (50.2) \\
    \noalign{\hrule height 1.5pt}
  \end{tabular}}
  \caption{Comparison of our models (\textsc{UniversalRules} and L-\textsc{WeaklySupervised}) with previous work on the
    English Penn Treebank. H1 and H2 are two heuristics used in \citet{rasooli2012fast}.}
  \label{tab:wsj}
\end{table}

\begin{table*}[]
\centering
\scalebox{1.0}{\begin{tabular}{lccccc}
\noalign{\hrule height 1.5pt}
Model      & RF+H1+H2        & LC-DMV  & Conv-MST  & \textsc{L-WeaklySup} & \textsc{UnivRules} \\
\noalign{\hrule height 1pt}
\multicolumn{6}{c}{Length $\leq$ 15} \\                                      
\hline
Basque     & 49.0 (51.0) & 47.9 & 52.5 & \textbf{55.2} (56.0) & 52.9 (55.1) \\
Dutch      & 26.6 (31.9) & 35.5 & \textbf{43.4} & 38.7 (41.3) & 39.6 (40.2) \\
French     & 33.2 (37.5) & 52.1 & \textbf{61.6} & 56.6 (57.2) & 59.9 (61.6) \\
German     & 40.5 (44.0) & 51.9 & 54.4 & \textbf{59.7} (59.9) & 57.5 (59.4) \\
Italian    & 33.3 (38.9) & 73.1 & \textbf{73.2} & 58.5 (59.8) & 59.7 (62.3) \\
Polish     & 46.8 (59.7) & 66.2 & \textbf{66.7} & 61.8 (63.4) & 57.1 (59.3) \\
Portuguese & 35.7 (43.7) & \textbf{70.5} & 60.7 & 52.5 (54.1) & 52.7 (54.2) \\
Spanish    & 35.9 (38.3) & \textbf{65.5} & 61.6 & 55.8 (56.2) & 55.6 (56.8) \\
\hline
Average    & 37.6 (43.1) & 57.8 & \textbf{59.3} & 54.9 (56.0) & 54.4 (56.1)\\
\noalign{\hrule height 1pt}
\multicolumn{6}{c}{Length $\leq$ 40} \\                                      
\hline
Basque     & 45.4 (47.6) & 45.4 & 50.0 & \textbf{51.0} (51.7) & 48.9 (51.5)  \\
Dutch      & 23.1 (30.4) & 34.1 & \textbf{45.3} & 42.2 (44.8) & 42.5 (44.3) \\
French     & 27.3 (30.7) & 48.6 & \textbf{62.0} & 46.4 (47.5) & 55.4 (56.3) \\
German     & 32.5 (37.0) & 50.5 & 51.4 & \textbf{55.6} (56.3) & 54.2 (56.3) \\
Italian    & 27.7 (33.0) & \textbf{71.1} & 69.1 & 54.1 (55.6) & 55.7 (58.7) \\
Polish     & 43.3 (46.0) & \textbf{63.7} & 63.4 & 57.3 (59.4) & 51.7 (52.8) \\
Portuguese & 28.8 (35.9) & \textbf{67.2} & 57.9 & 44.6 (48.6) & 45.3 (46.5) \\
Spanish    & 26.9 (28.8) & \textbf{61.9} & \textbf{61.9} & 50.8 (54.0) & 52.4 (53.9) \\
\hline
Average    & 31.9 (36.2) & 55.3 & \textbf{57.6} & 50.3 (52.2)  & 50.8 (52.5)\\
\noalign{\hrule height 1pt}
\end{tabular}}
\caption{Evaluation on eight languages of the UD treebank with test
  sentences of length $\leq$ 15 and length $\leq$ 40.}
\label{tab:ud}
\end{table*}

\paragraph{English Penn Treebank}
We compare our unsupervised \textsc{UniversalRules} model and its
\textsc{WeaklySupervised} variant with (1)~the state-of-the-art
unsupervised transition-based parser of
\citet{rasooli2012fast},\footnote{Since we used different
  preprocessing, we re-implemented their model for a fair comparison.}
denoted as RF, and (2)~two state-of-the-art unsupervised graph-based
parsers with universal linguistic rules: Convex-MST
\citep{grave2015convex} and HDP-DEP \citep{naseem2010using}. Both of
these are not transition-based, and thus not directly comparable to
our approach, but are useful for reference.

The parser of \citet{rasooli2012fast} is unlexicalized and
count-based.  To reach the best performance, the authors employed
``baby steps'' (i.e.,~they start training on short sentences and
gradually add longer sentences \citep{spitkovsky2009baby}), as well
as two heuristics called H1 and H2.  
H1 involves multiplying the probability of the last verb reduction in a 
sentence by $10^{-10}$. 
H2 involves multiplying each \textit{Noun $\to$ Verb}, \textit{Adjective
  $\to$ Verb}, and \textit{Adjective $\to$ Noun} rule by $0.1$. 
These heuristics seem fairly ad-hoc; they presumably bias the probability
estimates towards more linguistically plausible values.

As the results in Table~\ref{tab:wsj} show, our
\textsc{UniversalRules} model outperforms RF on both WSJ-10 and full
WSJ, achieving a new state of the art for transition-based dependency
grammar induction. The RF model does not use universal rules, but its
linguistic heuristics play a similar role, which makes our comparison
fair.  Note that our L-\textsc{Weakly\-Supervised} model achieves a
further improvement over \textsc{UniversalRules}, making it comparable
with Convex-MST and HDP-DEP, demonstrating the potential of the
neural, transition-based dependency grammar induction approach, which
should be even clearer on large datasets.

\paragraph{Universal Dependency Treebank}
Our multilingual experiments use the UD treebank. There we evaluate
the two models that perform the best on the WSJ: the unlexicalized
\textsc{UniversalRule} model and lexicalized
L-\textsc{WeaklySupervised} model. We use the same hyperparameters as
in the WSJ experiments.  Again, we mainly compare our models with the
transition-based model RF (with heuristics H1 and H2), but we also
include the graph-based Convex-MST and LC-DMV models for reference.

Table~\ref{tab:ud} shows the UD treebank results.  It can be observed
that both \textsc{UniversalRules} and L-\textsc{WeaklySupervised}
significantly outperform the RF on both short and long sentences. The
improvement of average DDA is roughly 20\% on sentences of length
$\leq$ 40. This shows that although the heuristic approach employed by
\citet{rasooli2012fast} is useful for English, it does not generalize
well across languages, in contrast to our posterior-regularized
neural networks with universal rules.

\paragraph{Parsing Speed}
To highlight the advantage of our linear time complexity parser, we
compare both lexicalized and unlexicalized variants of our parser with
a representative DMV-based model (LC-DMV) in terms of parsing speed.
The results in Table~\ref{tab:speed} show that our unlexicalized
parser results in a 1.8-fold speed-up for short sentences (length
$\leq$ 15), and a speed-up of factor 16 for long sentences (full
length). And our parser does not lose much parsing speed even in a
lexicalized setting.

\begin{table}[t]
\centering
\begin{tabular}{l r r r} 
\noalign{\hrule height 1pt}
 Sentence length             & $\leq$15    & $\leq$40 & All \\
 \hline
 LC-DMV                 & 663   & 193  & 74  \\ 
 Our Unlexicalized      & 1192  & 1194 & 1191   \\
 Our Lexicalized        & 939   & 938  & 983 \\
\noalign{\hrule height 1pt}
\end{tabular}
\caption{Parsing speed (tokens per second) on the French
  UD Treebank with test sentences of various lengths. All
  experiments were conduct on the same CPU platform.}
\label{tab:speed}
\end{table}

\section{Related Work}
\label{sec:related}

In the family of graph-based models, besides LC-DMV, Convex-MST, and
HDP-DEP, a lot of work has focused on improving the DMV, such as
adding more types of valence \citep{headden2009improving}, training
with artificial negative examples \citep{smith2005guiding}, and
learning initial parameters from shorter sentences
\citep{spitkovsky2009baby}.  Among graph-based models, there is also
some work conceptually related to our
approach. \citet{jiang2017combining} combine a discriminative and a
generative unsupervised parser using dual
decomposition. \citet{cai2017crf} use CRF autoencoder for unsupervised
parsing.  In contrast to these two approaches, we use neural
variational inference to combine discriminative and generative models.

For transition-based models, \citet{daume2009unsupervised} introduces
a structure prediction approach and \citet{rasooli2012fast} propose a
model with simple features based on this approach.  Recently,
\citet{shi-huang-lee:2017:EMNLP2017} and
\citet{gmezrodrguez-shi-lee:2018:Long} show that practical dynamic
programming-like global inference is possible for transition-based
systems using a minimal set of bidirectional LSTM features. These
models achieve competitive performance for projective or
non-projective supervised dependency parsing but have not been applied
to unsupervised parsing.


\section{Conclusions}
\label{sec:conc}

In this work, we propose a neural variational transition-based model for dependency grammar induction.  
The model consists of a generative RNNG for generation, and a discriminative RNNG for parsing and inference.  
We train the model on unlablled corpora with an integration of neural variational inference, posterior regularization and
variance reduction techniques.
This allows us to use a small amount of universal linguistic rules as prior knowledge to regularize the latent space.
We show that it is straightforward to integrate weak
supervision into our model in the form of rule expectations.
Our parser obtains a new state
of the art for unsupervised transition-based dependency parsing, with linear time complexity and
 significantly faster parsing speed compared to graph-based models.

In future, we plan to conduct a larger-sclae of grammar induction experiment with our model.
We will also explore
better training and optimization techiniques for neural variational inference with discrete autoregressive latent variables.

\section{Acknowledgments}

We gratefully acknowledge the support of the Leverhulme Trust (award
IAF-2017-019 to FK).
We also thank Li Dong and Jiangming Liu at ILCC for fruitful discussions, 
Yong Jiang at ShanghaiTech for sharing preprocessed WSJ dataset, 
and the anonymous reviewers of AAAI-19 for the constructive comments.


\bibliography{aaai19}
\bibliographystyle{aaai}

\newpage
\section{Supplementary Material}
\label{sec:supmat}


\subsection{Derivation of the M-step for Revised EM}
\label{chap:M-step}

\paragraph{Original Score Function}
Under the posterior regularized EM framework, the original score function without a baseline should be defined as:
\begin{align*}
\log \frac{p_{\theta}(x, a)} {q(a)} = \log \frac{p_{\theta}(x, a)} {q_\omega(a|x) \gamma(a, x)}
\end{align*} 
But in practical training, we observed that $\gamma(a, x)$ will assign
large weights (larger than 1) to more likely parse trees and small 
weights (less than 1) to less likely parse trees. Thus $-\log \gamma(a, x)$ would
effectively reverse the direction of optimization, which could
dramatically mislead the learning process. To cope with this issue, we
simply define the score function for our revised EM algorithm as
Eq.~(7). We will show how this definition
will affect the loss function.
\begin{align*}
\begin{split}
 \mathbb{E}&_{q(a)}[\log\gamma(x,a)] \\
& = \sum_{a} q(a) \log\gamma(x,a) \\
& = \sum_{a} q_\omega(a | x) \gamma(x,a) \log \gamma(x,a) \\
& \leqslant \sum_{a} q_\omega(a | x) \gamma(x,a) \gamma(x,a) \\
& = \mathbb{E}_{q_\omega(a | x)}[\gamma^2(x,a)] \\
& = \mathrm{Var}_{q_\omega(a | x)}[\gamma(x,a)] + \mathbb{E}_{q_\omega(a | x)}[\gamma(x,a)]^2
\end{split}
\end{align*}
Since
\begin{align*}
\begin{split}
\mathbb{E}_{q_\omega(a | x)}[\gamma(x,a)] = \sum_{a} q_\omega(a | x) \gamma(x,a) = 1
\end{split}
\end{align*}
we have
\begin{align*}
\begin{split}
\mathbb{E}&_{q(a)}[\log \frac{p_{\theta}(x, a)} {q_\omega(a|x)}] \\
& = \mathbb{E}_{q(a)}[\log \frac{p_{\theta}(x, a)} {q_\omega(a|x) \gamma(a, x)} + \log \gamma(a, x)] \\
& = \mathbb{E}_{q(a)}[\log \frac{p_{\theta}(x, a)} {q(a)}] + \mathbb{E}_{q(a)}[\log \gamma(a,x)]\\
& \leqslant \mathcal{L}_x + \mathrm{Var}_{q_\omega(a | x)}[\gamma(x,a)] + 1.
\end{split}
\end{align*}
Thus, in theory, $\mathrm{Var}_{q_\omega(a | x)}[\gamma(x,a)]$ can be viewed as a regularization for posterior regularization.

\begin{algorithm}[!htbp]
    \SetAlgoLined
    \textbf{Parameters}: $\omega, \theta, \bm{\lambda}, \varepsilon, ||\cdot||_{\beta}, M$ \\
    \textbf{Constrained Feature Functions}: $\bm{\phi}(x,a)$ \\
    Initialization\;
    \While{not converged}{
        
        Sample $a^{(m)} \sim q_{\omega}(a|x), 1 \leqslant m \leqslant M $\;
        PR Computation: \\
        $\:\:\:Z(\bm{\lambda}) \approx \frac{1}{M} \sum_m \exp(-\bm{\lambda}^T\bm{\phi}(x, a^{(m)})),$
        $\gamma(x,a^{(m)}) = \frac{1}{Z(\bm{\lambda})} \exp(-\bm{\lambda}^{T}\bm{\phi}(x, a^{(m)}))$; \\
        Update parameters in mini-batch: \\
        $\:\:$Update $\theta$ w.r.t. its gradient $\frac{1}{M} \sum_m \gamma(x, a^{(m)}) \frac{\partial \log p_\theta (x, a^{(m)})}{\partial \theta},$ \\
        $\:\:$Update $\omega$ w.r.t. its gradient $\frac{1}{M} \sum_m \gamma(x, a^{(m)}) \frac{\partial \log q_{\omega}(a|x)}{\partial \omega}$,  \\
        $\:\:$Update $\bm{\lambda}$ to optimize $\max_{\bm{\lambda} \geqslant 0} ~~ -\bm{b}^T\bm{\lambda} - \log Z(\bm{\lambda}) - \varepsilon ||\bm{\lambda}||_{\beta^*}$ \small{(with projected gradient descent algorithm)}.
    }
    \caption{Pretraining for Variational Inference Dependency Parser.}
    \label{algo:pre}
\end{algorithm}

\paragraph{Parameter Updating}
In the revised EM algorithm, the parameters of both encoder and
decoder should be updated under the distribution of $q(a)$ rather than
$p_\omega(a|x)$. Since $q(a) = \gamma(x,a) p_\omega(a|x)$, the
gradient for the parameter of the encoder via MC sampling will be:
\begin{align*}
    &\frac{\partial \mathcal{L}_x}{\partial \omega} 
    = \frac{1}{M} \sum_m  \gamma(x, a^{(m)}) l(x, a^{(m)}) \frac{\partial \log q (a^{(m)})}{\partial \omega} \\
    &= \frac{1}{M} \sum_m  \gamma(x, a^{(m)}) l(x, a^{(m)}) \frac{\partial \log p_{\omega}(a^{(m)}|x) } {\partial \omega}
\end{align*}
For the decoder, to boost the performance, we also use the score function for optimization:
\begin{align*}
    &\frac{\partial \mathcal{L}_x}{\partial \theta} \\
    &= \frac{1}{M} \sum_m  \gamma(x, a^{(m)}) l(x, a^{(m)}) \frac{\partial \log p_\theta (x, a^{(m)})}{\partial \theta}
\end{align*}

\begin{algorithm}[!htbp]
    
    \SetAlgoLined
    \textbf{Parameters}: $\omega, \theta, \bm{\lambda}, \varepsilon, ||\cdot||_{\beta}, M$ \\
    \textbf{Constrained Feature Functions}: $\bm{\phi}(x,a)$ \\
    \textbf{Critic}: $l_{\textsc{critic}}(x,a)$ \\
    Initialization\;
    \While{not converged}{
        \textbf{E$^\prime$-step}: \\
        Sample $a^{(m)} \sim q_{\omega}(a|x), 1 \leqslant m \leqslant M $\;
        PR Computation: \\
        $\:\:\:Z(\bm{\lambda}) \approx \frac{1}{M} \sum_m \exp(-\bm{\lambda}^T\bm{\phi}(x, a^{(m)})),$
        $\gamma(x,a^{(m)}) = \frac{1}{Z(\bm{\lambda})} \exp(-\bm{\lambda}^{T}\bm{\phi}(x, a^{(m)}))$, \\
        $q(a^{(m)}) = q_{\omega}(a^{(m)}|x) \gamma(x,a^{(m)})$; \\
        Compute $l_{\textsc{critic}}(x,a) $ for specific critic;\\
        \textbf{M-step}: \\
        Update parameters in mini-batch: \\
        $\:\:$Update $\theta$ w.r.t. its gradient $\frac{1}{M} \sum_m \gamma(x, a^{(m)}) \frac{\partial \log p_\theta (x, a^{(m)})}{\partial \theta},$ \\
        $\:\:$Update $\omega$ w.r.t. its gradient $\frac{1}{M} \sum_m \gamma(x, a^{(m)}) \frac{\partial \log q_{\omega}(a|x)}{\partial \omega}$,  \\
        $\:\:$Update $\bm{\lambda}$ to optimize $\max_{\bm{\lambda} \geqslant 0} ~~ -\bm{b}^T\bm{\lambda} - \log Z(\bm{\lambda}) - \varepsilon ||\bm{\lambda}||_{\beta^*}$ \small{(with projected gradient descent algorithm)}.
    }
    
    \caption{Revised EM Algorithm for Variational Inference Dependency Parser.}
    \label{algo:em}
\end{algorithm}

\subsection{Linguistic Rules}
\label{chap:rules}

Table \ref{table:WSJ PR rules} and \ref{table:UD PR rules} present the universal linguistic rules used for WSJ and the Universal Dependency Treebank respectively.

\begin{table}[!htbp]
  \centering
  \begin{tabular}{| l | l |}
    \hline
    Root $\to$ Auxiliary    & Noun $\to$ Adjective \\
    Root $\to$ Verb         & Noun $\to$ Article \\ \cline{1-1}
    Verb $\to$ Noun         & Noun $\to$ Noun \\
    Verb $\to$ Pronoun      & Noun $\to$ Numeral \\ \cline{2-2} 
    Verb $\to$ Adverb       & Preposition $\to$ Noun \\ \cline{2-2} 
    Verb $\to$ Verb         & Adjective $\to$ Adverb \\ \hline
    Auxiliary $\to$ Verb    &  \\ \hline
  \end{tabular}
  \caption{Universal dependency rules for WSJ \citep{naseem2010using}.}
  \label{table:WSJ PR rules}
\end{table}

\begin{table}[!htbp]
  \centering
  \begin{tabular}{| l | l |}
    \hline
    ROOT $\to$ VERB         & NOUN $\to$ ADJ \\
    ROOT $\to$ NOUN         & NOUN $\to$ DET \\ \cline{1-1}
    VERB $\to$ NOUN         & NOUN $\to$ NOUN \\
    VERB $\to$ ADV          & NOUN $\to$ NUM \\  
    VERB $\to$ VERB         & NOUN $\to$ CONJ \\ 
    VERB $\to$ AUX          & NOUN $\to$ ADP \\ \hline
    ADJ $\to$ ADV           &  \\ \hline
  \end{tabular}
  \caption{Universal dependency rules for the Universal Dependency Treebank \citet{noji2016using}.}
  \label{table:UD PR rules}
\end{table}

\subsection{Experimental Details}
\label{chap:exp details}

Algorithm \ref{algo:pre} and \ref{algo:em} provide the outlines of our pretraining and revised EM algorithms respectively. To avoid starting training from arbitrarily initialized encoder and decoder, we pretrain our encoder and decoder separately via posterior regularization, where $\gamma(x, a)$ serves the reward for the REINFORCE.

\subsection{Model Configuration}
\label{chap:model conf}

\paragraph{Encoder and Decoder}
We follow the model configuration in \citet{cheng2017generative} to build the encoder and decoder by using Stack-LSTMs \citep{dyer2015transition}. The differences are (1) we use neither the parent non-terminal embedding nor the action history embedding for both the decoder and encoder; (2) we do not use the adaptive buffer embedding for the encoder. 

\vspace{-1em}
\paragraph{RL-BL}
For the baseline ($b(x)+b$) in RL-BL, we first pretrain a LSTM language model. We use word embeddings of size 100, 2 layer LSTM with 100 hidden size, and tie weights of output classifiers and word embeddings. During training the RL-BL, we fix the LSTM language model and rescale and shift the output $\log p(x)$ to fit the ELBO of the given sentence as
\begin{align*}
b(x) + b = \alpha \log p(x) + \tau
\end{align*}

\paragraph{Hyper-Parameters and Optimization}
\begin{table}[tb]
  \centering
  \begin{tabular}{l c}
    \hline
    word embeddings dimensions  & 80 \\
    PoS embeddings dimensions   & 80 \\
    Encoder LSTM dimensions     & 64 \\
    Decoder LSTM dimensions     & 64 \\
    LSTM layer                  & 1  \\
    Encoder dropout             & 0.5 \\
    Decoder dropout             & 0.5 \\
    Learning rate               & 0.01 \\
    gradient clip               & 0.25 \\
    gradient clip (for pretraining)& 0.5 \\
    $\ell_2$ regularization     & 1e-4 \\
    $\varepsilon$ in~Eq.~(8)    & 0.1 \\
    number of MC samples        & 20 \\
    \hline
  \end{tabular}
  \caption{Hyperparameters.}
  \label{tab:hyper}
\end{table}

In all experiments, both PoS tag and word embeddings are used in our lexicalized models while only PoS tag embeddings are used in our unlexicalized models. Table~\ref{tab:hyper} presents the hyperparameter settings. For pretrained word embeddings, we project them into lower dimension (word embedding dimensions). We select $\varepsilon$ in~Eq.~(\ref{eq:PR set}) via a grid search on WSJ-10 development set. And we use Glorot for parameter initialization, and Adagrad for optimization except for posterior regularization.

\end{document}